\documentclass[journal,twoside,web]{ieeecolor}
\usepackage[table]{xcolor}
\usepackage{generic}
\usepackage{amsmath,amssymb,amsfonts}
\usepackage{algorithmic}
\usepackage{graphicx}
\usepackage{algorithm,algorithmic}
\usepackage{cite}    
\makeatletter
\let\NAT@parse\undefined
\makeatother
\usepackage{hyperref}

\usepackage{textcomp}
\usepackage{booktabs}
\usepackage{diagbox}
\usepackage{pifont}
\newcommand{\cmark}{\ding{51}}  
\newcommand{\xmark}{\ding{55}}  
\newcommand{\tablestyle}[2]{\setlength{\tabcolsep}{#1}\renewcommand{\arraystretch}{#2}\centering\footnotesize}
\newcommand{\thename}[0]{DeltaMIL}

\def\BibTeX{{\rm B\kern-.05em{\sc i\kern-.025em b}\kern-.08em
    T\kern-.1667em\lower.7ex\hbox{E}\kern-.125emX}}
\begin{document}
\title{DeltaMIL: Gated Memory Integration for Efficient and Discriminative Whole Slide Image Analysis}
\author{Yueting Zhu, Yuehao Song, Shuai Zhang, Wenyu Liu, \IEEEmembership{Senior Member, IEEE}, and Xinggang Wang, \IEEEmembership{Member, IEEE}
\thanks{Yueting Zhu, Yuehao Song, Shuai Zhang, Wenyu Liu, and Xinggang Wang are with the School of Electronic Information and Communications, Huazhong University of Science and Technology, Wuhan, 430074, China. (e-mail: yuetingzhu@hust.edu.cn; yhaosong@hust.edu.cn; shuaizhang@hust.edu.cn; liuwy@hust.edu.cn; xgwang@hust.edu.cn). }
}

\maketitle

\begin{abstract}
Whole Slide Images (WSIs) are typically analyzed using multiple instance learning (MIL) methods. 
However, the scale and heterogeneity of WSIs generate highly redundant and dispersed information, making it difficult to identify and integrate discriminative signals.
Existing MIL methods either fail to discard uninformative cues effectively or have limited ability to consolidate relevant features from multiple patches, which restricts their performance on large and heterogeneous WSIs.
To address this issue, we propose \thename{}, a novel MIL framework that explicitly selects semantically relevant regions and integrates the discriminative information from WSIs.
Our method leverages the gated delta rule to efficiently filter and integrate information through a block combining forgetting and memory mechanisms.
The delta mechanism dynamically updates the memory by removing old values and inserting new ones according to their correlation with the current patch.
The gating mechanism further enables rapid forgetting of irrelevant signals.
Additionally, \thename{} integrates a complementary local pattern mixing mechanism to retain fine-grained pathological locality. 
Our design enhances the extraction of meaningful cues and suppresses redundant or noisy information, which improves the model’s robustness and discriminative power.
Experiments demonstrate that \thename{} achieves state-of-the-art performance. Specifically, for survival prediction, \thename{} improves performance by 3.69\% using ResNet-50 features and 2.36\% using UNI features. For slide-level classification, it increases accuracy by 3.09\% with ResNet-50 features and 3.75\% with UNI features. These results demonstrate the strong and consistent performance of \thename{} across diverse WSI tasks.
\end{abstract}

\begin{IEEEkeywords}
Whole slide image, Multiple instance learning, Gated delta rule, Memory update mechanism
\end{IEEEkeywords}

\section{Introduction}
\label{sec:introduction}
\IEEEPARstart{W}{hole} Slide Image (WSI) plays a central role in computational pathology, providing ultra-high-resolution images for slide-level diagnosis and prognosis~\cite{pathsurvey, naturedata}. 
Due to the large size of WSIs, Multiple Instance Learning has become the standard paradigm~\cite{mil}.
In MIL, a slide is divided into small patches, which are treated as instances, and represented as a bag of aggregated patch embeddings~\cite{clam,deepmil,mmil,uni,coach,titan} for training.
Inspired by the significant impact of the Transformer architecture~\cite{transformer} on various downstream tasks~\cite{vitdet,vitpose,vitgaze,vitmatte}, Transformer-based MIL methods~\cite{clam,abmil,dsmil,transmil} are introduced to more effectively integrate information across patches and capture complex inter-patch relationships.
Linear or subquadratic models~\cite{s4mil,mambamil,mamba2mil,2dmamba,gmmamba,m3amba} further enhance feature aggregation, enabling strong performance on tasks such as slide-level classification and survival prediction.

\begin{figure}[!t]
\centering
\includegraphics[width=0.98\columnwidth]{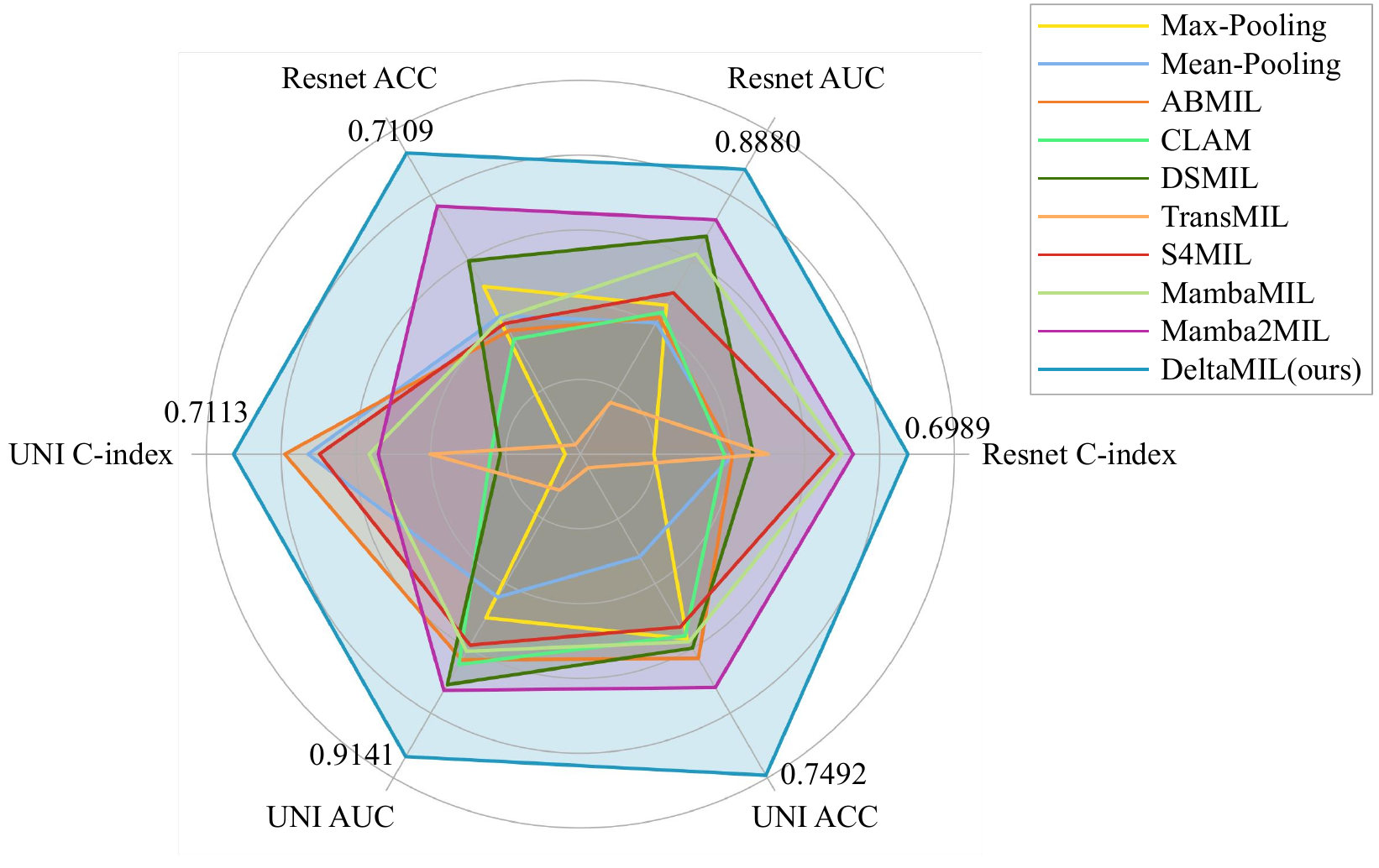}
\vspace{-5pt}\caption{Comprehensive comparison across diverse whole-slide image (WSI) analysis tasks, showing strong and well-balanced performance.}\vspace{-12pt} 
\label{fig:radar}
\end{figure}

\begin{figure}[!t]
\centering
\includegraphics[width=0.98\columnwidth]{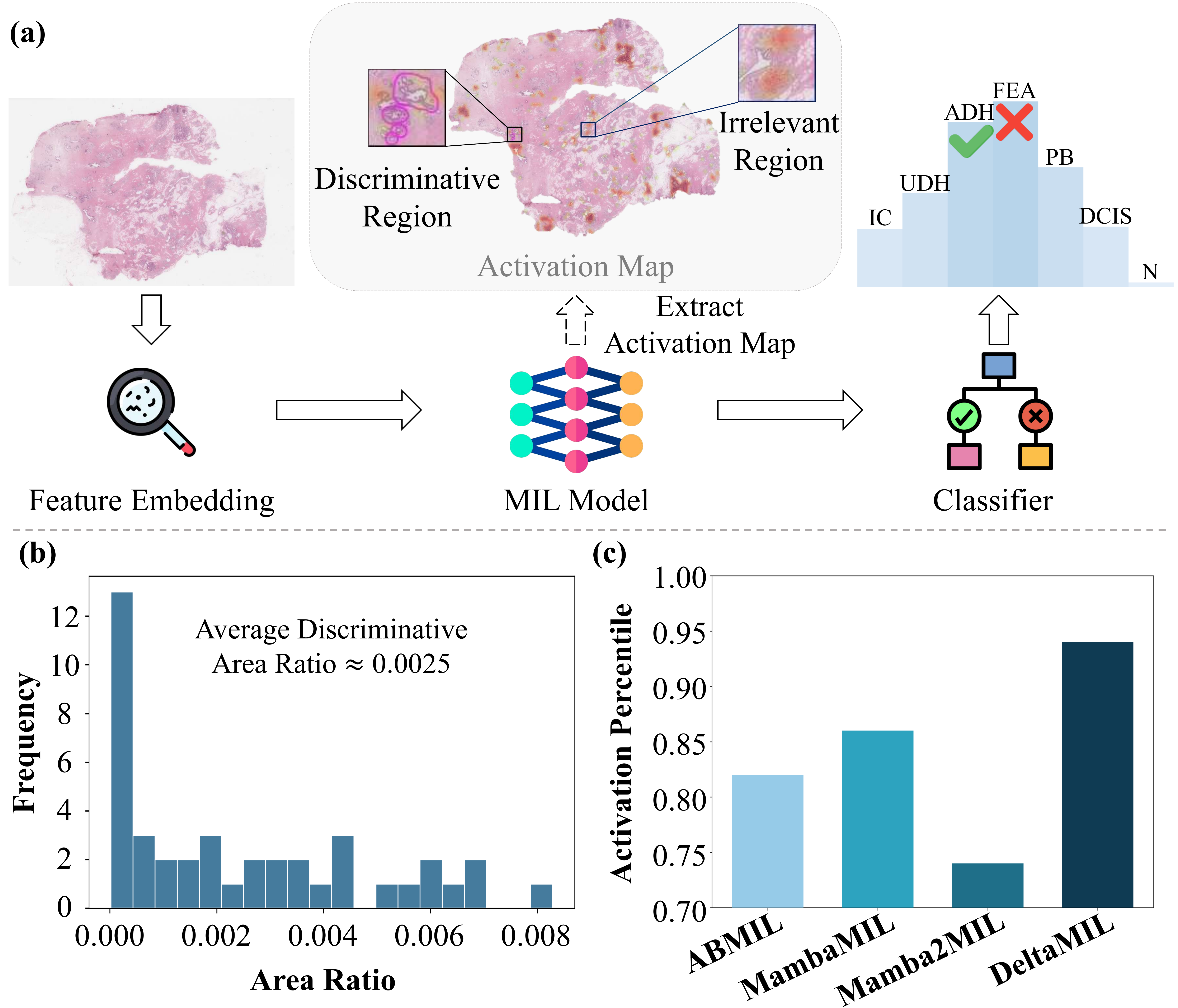}
\vspace{-5pt}\caption{Irrelevant regions interfere with the MIL models’ analysis on WSIs.
(a) Existing MIL models allocate high activation to irrelevant regions during activation map extraction, which finally results in prediction errors.
(b) Area ratio distribution of annotated discriminative regions in the BRACS~\cite{bracs} test fold. The area of discriminative regions constitutes only a tiny portion of the whole slide ($\approx 0.25\%$).
(c) Average percentiles of the activation scores within the annotated discriminative regions of different MIL models. \thename{} assigns the highest activation score.}
\label{fig:teaser}\vspace{-10pt}
\end{figure}

Despite the advances of the MIL methods, it remains a significant challenge that the discriminative regions occupy a small proportion of the WSI. The model's performance is interfered with by the large portions of the irrelevant tissues.
Accurate identification of these critical regions is essential for reliable diagnosis, yet the extreme sparsity of informative areas makes it difficult for models to focus effectively.
Fig.~\ref{fig:teaser}(b) illustrates that the discriminative regions occupy only a tiny fraction($\approx 0.25\%$) of the WSI and lead to a huge class imbalance of the instance bag.
When MIL models extract and aggregate patch-level features from the WSI, they often fail to assign sufficient attention to discriminative regions.
This diluted feature representation finally leads to the wrong predictions (Fig.~\ref{fig:teaser}(a)).
These challenges motivate the development of methods that can selectively focus on the discriminative regions while disregarding irrelevant regions.

Built upon these observations, we propose \thename{}, a framework that incorporates a forget–update memory mechanism to suppress irrelevant information and strengthen discriminative features.
\thename{} adapts the gated delta update principle~\cite{gateddelta} to the MIL setting to selectively maintain and refresh global memory as each patch is processed.
Specifically, the delta component dynamically updates the memory by removing outdated values and inserting new ones based on their correlation with the current patch, enabling the model to incrementally integrate relevant information while focusing on diagnostically important patches.
The gating component determines the relevance of patches dynamically with each update and selectively regulates the information to be forgotten, which filters out non-informative signals and preserves clinically meaningful information.
To effectively capture local details while maintaining the overall contextual information, our approach integrates local features with global context through weighted fusion. By learning adaptive weights, the model can dynamically balance the contribution of local and global information.
With the above design, our approach effectively suppresses irrelevant information and enhances discriminative features, as reflected by the average activation percentiles shown in Fig.~\ref{fig:teaser}(c).

Extensive experiments demonstrate the effectiveness of \thename{} in handling whole-slide images by reducing redundant information and improving the integration of dispersed signals, which leads to consistent performance gains across tasks. For survival prediction, we evaluated the method on eight datasets and achieved state-of-the-art results, with improvements of 3.69\% using ResNet-50~\cite{resnet} features and 2.36\% using UNI~\cite{uni} features. For slide-level classification, tests on two datasets show that \thename{} also outperforms existing approaches, achieving gains of 1.79\% AUC and 3.09\% ACC with ResNet-50~\cite{resnet} features, and 2.05\% AUC and 3.75\% ACC with UNI~\cite{uni} features.
These results demonstrate that \thename{} effectively mitigates the impact of redundant and scattered information in WSIs, leading to more robust and accurate slide-level predictions across diverse datasets.

The main contributions of this paper are as follows:
\begin{itemize}
    \item We uncover a previously overlooked but widespread problem: task-irrelevant regions often occupy a substantial portion of the representation capacity in the feature maps, interfering with the learning of discriminative cues and consequently hindering overall model performance.
    \item We propose \thename{}, a framework that leverages the gated delta rule to explicitly emphasize informative regions while suppressing irrelevant ones, and further fuses fine-grained local information with global context to enhance overall model performance.
    \item Extensive experiments on multiple benchmarks demonstrate that \thename{} achieves state-of-the-art performance for both slide-level classification (3.09\% improvement in ACC) and survival analysis (3.69\% improvement in C-index).
\end{itemize}

\section{RELATED WORK}
\subsection{MIL for Whole Slide Images}
Multi-instance learning methods~\cite{mil} first split the whole slide into multiple patches and then extract patch-level features using pre-trained foundation models~\cite{resnet,uni,coach,evax,titan}.
Aggregating patch-level features into meaningful slide-level representations is a core challenge in WSI analysis. Early approaches rely on simple strategies such as Max-Pooling and Mean-Pooling~\cite{deepmil}, while they often neglect the varying importance of different regions. To address this limitation, attention-based MIL methods~\cite{clam,abmil} leverage an attention mechanism~\cite{transformer} to adaptively weigh the contributions of the regions.

Subsequent works~\cite{dsmil,transmil,dtfd,mmil,policy, oodml,sdmil} extend attention-based MIL by introducing stronger feature aggregation modules and more expressive inter-patch dependency modeling to enhance slide-level reasoning. However, these methods typically rely on multi-stage pipelines and focus primarily on a subset of selected instances, which limits their ability to capture holistic contextual information.

Recently, state space models~\cite{S4,mamba,vim, m3amba, gmmamba} have emerged to efficiently handle the extremely long sequences of patches in WSIs. S4MIL~\cite{s4mil} applies structured state space models to compress sequences into memory units while preserving long-range dependencies. MambaMIL~\cite{mamba} leverages the Mamba sequential model and sequence reordering to capture comprehensive relationships among instances.
Mamba2MIL~\cite{mamba2mil} extends this approach by introducing weighted feature selection and sequence transformation to enable effective feature fusion.
2DMamba~\cite{2dmamba} generalizes the Mamba framework to 2D spatial structures using a hardware-aware operator to model wide 2D contexts efficiently.
Despite their strengths in capturing global dependencies, these methods often fail to filter uninformative or redundant information, which limits their ability to fully leverage the rich, heterogeneous patterns present in WSIs.

\begin{figure*}[!t]
\centering
\includegraphics[width=0.98\textwidth]{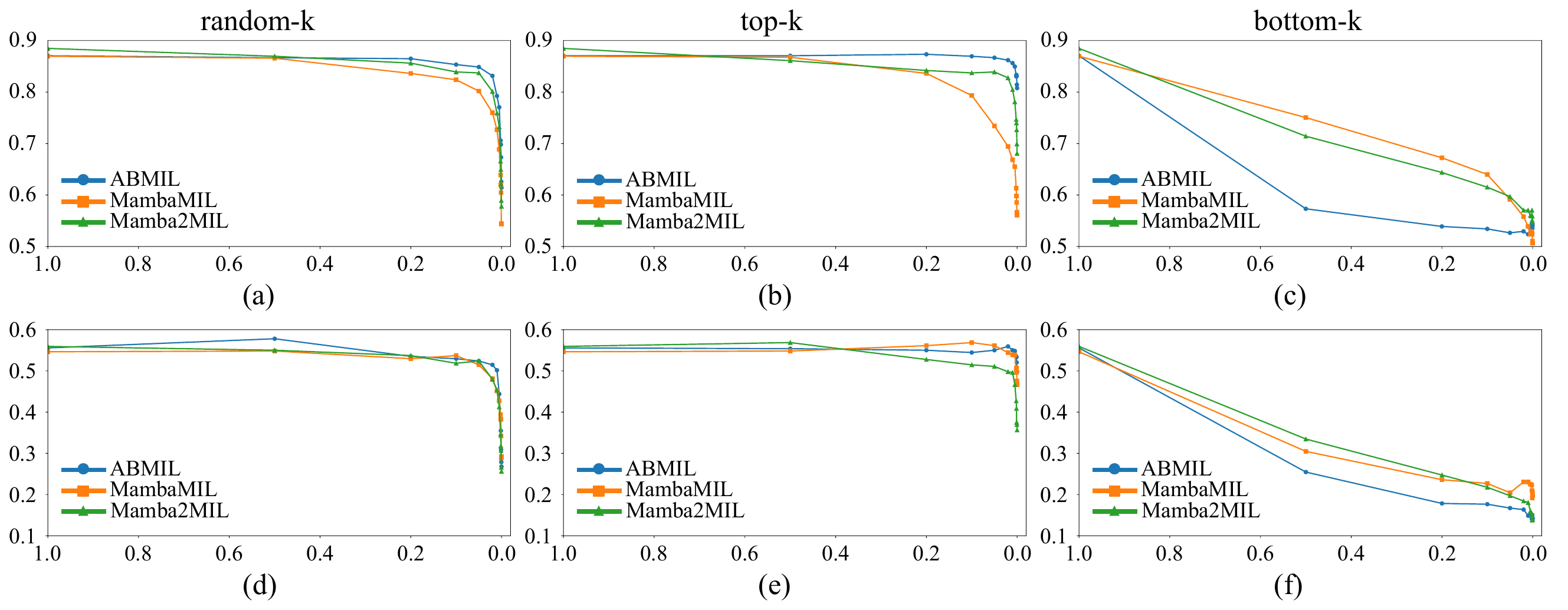}
\vspace{-3pt}\caption{Accuracy variation with respect to the retained patch ratio.
This figure presents the changes in model ACC on the NSCLC dataset (panels (a)–(c)) and the BRACS~\cite{bracs} dataset (panels (d)–(f)) as the proportion of retained patches decreases. Each set of panels corresponds to the three sampling strategies, Random-k, Top-k, and Bottom-k, applied to ABMIL~\cite{abmil}, MambaMIL~\cite{mambamil}, and Mamba2MIL~\cite{mamba2mil}. The pronounced differences among the three strategies indicate that the model’s decision-making relies primarily on a tiny subset of highly informative patches, while most patches are irrelevant to the final prediction, demonstrating the inherent redundancy in pathology images.}\vspace{-7pt}
\label{fig:motivation}
\end{figure*}

\subsection{Efficient Sequence Modeling}
To overcome the $O(N^2)$ complexity bottleneck of Transformer attention~\cite{transformer}, recent sequence modeling research~\cite{lstm,la,griffin,retnet} focuses on integrating structured recurrence and data-dependent gating to achieve subquadratic complexity architectures.
State space models with recursive structures leverage structured matrices to achieve stable long-range memory and improve model scalability. S4~\cite{S4} employs a HiPPO-structured matrix, S5~\cite{S5} and LRU~\cite{LRU} adopt simplified diagonal structures, and Mamba~\cite{mamba} introduces a data-dependent selection mechanism that dynamically adjusts SSM parameters, enabling highly selective, content-aware behavior. RWKV~\cite{RWKV} achieves attention-like behavior through a compact combination of time-mixing for temporal aggregation and channel-mixing for position-wise transformation. GLA~\cite{GLA} and HGRN~\cite{HGRN} further extend linear recurrent architectures by incorporating gating mechanisms that enhance their expressiveness and stability.
However, despite their efficiency, most existing approaches overemphasize global dependencies and underappreciate local variations, making them less effective at discriminating informative patches from the vast amount of task-irrelevant tissue in whole-slide images.

DeltaNet~\cite{deltanet} introduces the delta update rule to selectively update key-value memory, efficiently maintaining global context over very long sequences.
Gated DeltaNet~\cite{gateddelta,infinitevl} further improves the update efficiency by incorporating a learnable gating mechanism into the delta rule, enabling selective memory refreshing and more efficient global modeling.
However, Gated DeltaNet still focuses primarily on global dependencies and does not explicitly enhance fine-grained local modeling.

\begin{figure*}[!t]
\centering
\includegraphics[width=0.98\textwidth]{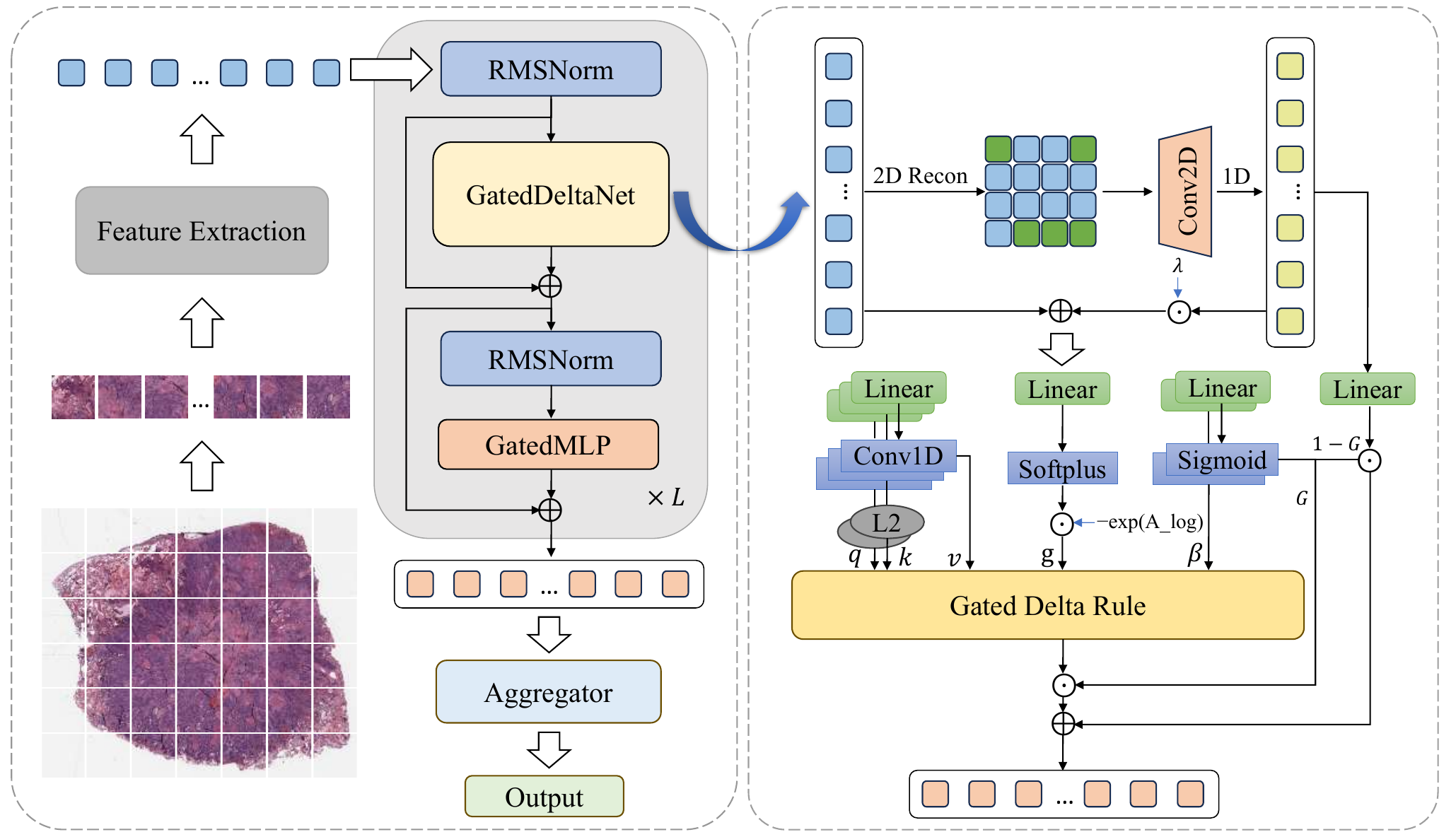}
\vspace{-3pt}\caption{Overall framework of the proposed model.
The left part shows the overall workflow: WSIs are processed by a feature extractor and then pass through $L$ stacked Gated DeltaNet blocks, followed by an aggregator to produce slide-level predictions. The right part details the Gated DeltaNet module, which enhances the original Gated DeltaNet by integrating local features reconstructed from the 2D spatial layout. This design enables joint modeling of long-range dependencies and fine-grained spatial patterns in histopathology data.}
\label{fig:overall}\vspace{-10pt}
\end{figure*}

\section{METHOD}
\subsection{Motivation}
In WSI analysis, only the lesion contributed to the definitive diagnosis, while most regions are irrelevant to the final predictions.
This phenomenon suggests that the model primarily relies on a small amount of critical information, and the presence of many other patches has minimal effect on the overall prediction.

To quantify this phenomenon, we designed a series of experiments. First, for each method, we compute the attention scores of each patch:
\begin{equation}
\alpha_i = \text{Attention}(x_i; \theta_\text{method}), \quad i = 1, \dots, N,
\end{equation}
where $x_i$ denotes the $i$-th patch of the input image, $\theta_\text{method}$ represents the trained model parameters for each method, and $\alpha_i$ is the attention value of the patch.
Based on these attention scores, we define three patch sampling strategies: random-k, top-k, and bottom-k.
Let $P \subset \{1, \dots, N\}$ denote the set of selected patches:
\begin{align}
P_\text{random-k} &\subset \{1,\dots,N\}, \ |P_\text{random-k}| = k,\\
P_\text{top-k} &= \{ i \mid \alpha_i \ge \alpha_{(k)} \}, \\
P_\text{bottom-k} &= \{ i \mid \alpha_i \le \alpha_{(N-k+1)} \}, 
\end{align}
where $\alpha_{(k)}$ denotes the $k$-th largest attention value, and $\alpha_{(N-k+1)}$ denotes the $k$-th smallest attention value.
Here, random-k serves as a baseline, while top-k and bottom-k evaluate the significance of irrelevant and discriminative patches, respectively.
Using each sampling strategy, we select a subset of patches $P$ and then feed them back into the original model to recompute the classification prediction:
\begin{equation}
\hat{y} = f_{\theta_\text{method}}(\{ x_i \}_{i \in P}).
\end{equation}

We vary the retained patch ratio $k/N$ and repeat the experiment above to observe the tendency of classification accuracy (ACC) to quantify the model's reliance on patches.
The tendency curves are shown in Fig.~\ref{fig:motivation}. We evaluate three patch sampling strategies on the NSCLC and BRACS~\cite{bracs} datasets using three models: ABMIL~\cite{abmil}, MambaMIL~\cite{mambamil}, and Mamba2MIL~\cite{mamba2mil}.
Based on these results, we observe that
\begin{itemize}
    \item \textbf{Top-k}: When retaining only about 5\%–10\% of high-attention patches, the model’s ACC remains close to the level achieved using all patches, suggesting that most patches are irrelevant to the model prediction.
    \item \textbf{Bottom-k}: When removing 50\% of the most informative patches, the model’s ACC drops sharply from 0.87 to 0.57, indicating that discarding high-value regions substantially undermines the model’s predictive ability.
\end{itemize}

The experimental results demonstrate that a small number of high-attention patches largely determine model predictions, while most patches contribute little, highlighting the problem of efficiently leveraging key information and avoiding the influence of uninformative patches.

\subsection{Overall Architecture}
As illustrated in Fig.~\ref{fig:overall}, our framework takes a whole-slide image as input and produces a compact slide-level representation for downstream prediction.
Each WSI is first divided into non-overlapping patches.
We obtain patch-level features together with their spatial coordinate using a pretrained feature extractor~\cite{resnet,uni}.
To alleviate the effect of the blank background regions, we discard these regions and preserve only foreground features, resulting in a sequence of patch features \( X = \{ x_1, x_2, \dots, x_N \} \), where each \( x_i \in \mathbb{R}^c \). The variables \( N \) and \( c \) denote the number of informative patches and the feature dimension, respectively.
Each patch feature \( x_i \) is mapped to a \( d \)-dimensional embedding through a fully connected layer, producing the embedded sequence \( Z = \{ z_1, z_2, \dots, z_N \} \) where \( z_i \in \mathbb{R}^d \). This sequence is then passed through a stack of \( L \) hierarchical Gated DeltaNet (GDN) blocks, denoted as \( Z_{\text{out}} = \mathrm{GDN}^{(L)}(Z) \), where each block refines patch representations by incorporating spatial coordinates and capturing both local and global dependencies. Finally, the output is aggregated into a compact slide-level representation \( R = \mathrm{Aggregate}(Z_{\text{out}}) \), which is fed into a classifier for prediction.

\subsection{GDNBlock}
The Gated DeltaNet Block(GDNBlock) is composed of the following components:  
1) a fused RMS normalization~\cite{rmsnorm} layer applied before attention to normalize the input features;
2) a Gated DeltaNet  module that combines the delta update rule with a local convolutional branch, enabling the block to capture both long-range dependencies and fine-grained local spatial features;
3) a second fused RMS normalization~\cite{rmsnorm} layer applied before the feed-forward network;  
4) a GatedMLP based on SwiGLU activation, which enhances channel-wise interactions through multiplicative gating.  

During the forward pass, residual connections and dropout are applied around both the attention and feed-forward modules, ensuring stable training and effective information flow. Overall, the block extends Gated DeltaNet with local spatial modeling, thereby improving its representation capability for patch-based histopathology data.

\subsection{Locality-Aware Gated Delta Rule}
As illustrated on the right of Fig.~\ref{fig:overall}, the 1D patch sequence is first reconstructed into a 2D layout based on spatial coordinates, forming a slide-level feature map. Positions corresponding to background or out-of-bound regions are filled with a learnable pad token. The feature map is then processed with a depthwise 2D convolution to capture local spatial context, and the resulting local features are mapped back to the 1D sequence and fused with the original embeddings through a learnable gate $\lambda$. Queries, keys, and values are then computed from the fused representation, where $\phi$ denotes a short convolutional preprocessing:
\begin{align}
Z_{\text{local}} &= \text{Conv2D}(\text{reconstruct2D}(Z, \text{coords})),\\
Z_{\text{local\_seq}} &= \text{extract2D}(Z_{\text{local}}, \text{coords}),\\
H &= Z + \tanh(\lambda) \cdot Z_{\text{local\_seq}},\\
Q =& \phi(W_q H), \quad 
K = \phi(W_k H), \quad 
V = \phi(W_v H).
\end{align}

\begin{figure}[t]
\centering
\includegraphics[width=0.98\columnwidth]{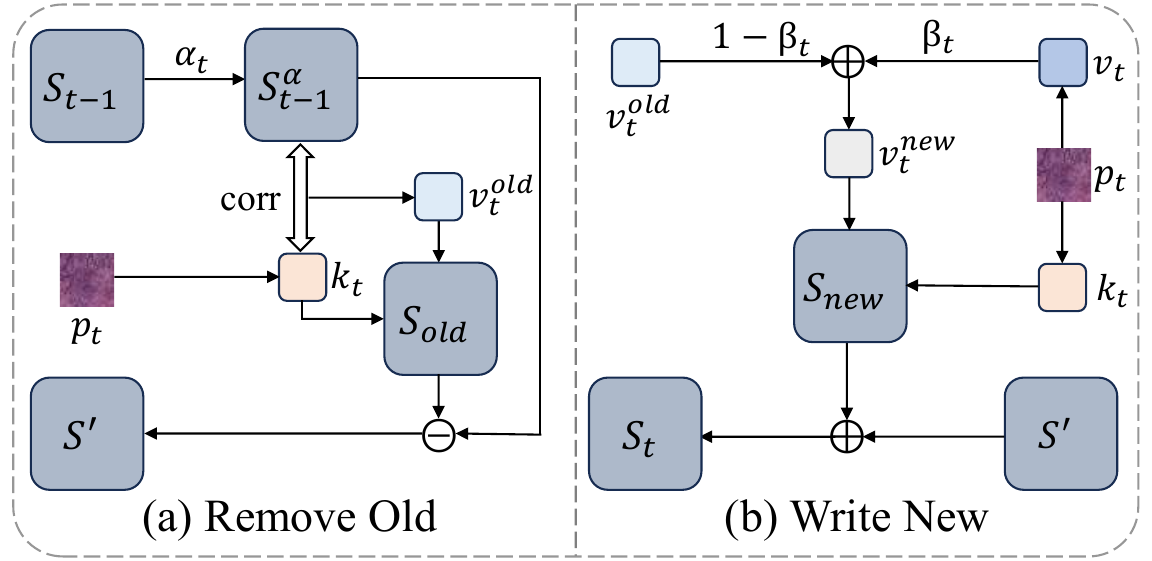}
\vspace{-3pt}\caption{Illustration of the Gated Delta rule, which consists of two operations: (a) Remove Old — discarding outdated information through a decay mechanism, and (b) Write New — updating with new information.}\vspace{-10pt}
\label{fig:rule}
\end{figure}

With the query, key, and value sequences obtained, the Gated DeltaNet~\cite{gateddelta} module performs a chunkwise gated delta update to maintain a compact global memory over long patch sequences. In this process, a learnable retention gate $\alpha_t$ regulates the preservation of previous memory, while an update gate $\beta_t$ controls the integration of new information, allowing the model to selectively retain salient context and efficiently discard stale or irrelevant features:
\begin{equation}
\beta_t = \sigma(W_\beta h_t), \qquad 
\alpha_t = \sigma(W_\alpha h_t).
\end{equation}

At each step, the global memory is first attenuated by a learnable retention gate $\alpha_t$, and the previous memory contribution corresponding to the current key is retrieved. The new value is then computed by blending the incoming information with the retained memory using the update gate $\beta_t$. Finally, the global memory matrix is updated according to the gated delta rule:
\begin{align}
v_t^{\text{old}} &= \alpha_t S_{t-1} k_t,\\
v_t^{\text{new}} &= \beta_t v_t + (1 - \beta_t) v_t^{\text{old}},\\
S_t &= \alpha_t S_{t-1} - v_t^{\text{old}} k_t^\top + v_t^{\text{new}} k_t^\top.
\end{align}

We further provide the insight behind the gate delta rule in Fig.~\ref{fig:rule}.
The gated delta update can be interpreted through two complementary operations: \textbf{removing old information} and \textbf{writing new information}. After computing the gated delta terms described above, the model first removes the previous contribution associated with the current patch's key $k_t$ by subtracting $v_t^{\text{old}} k_t^\top$ from the attenuated memory $\alpha_t S_{t-1}$. This operation effectively ``forgets'' the old representation of the WSI patch in the global memory. It then writes updated information back into the memory by adding $v_t^{\text{new}} k_t^\top$, which injects the newly computed features of the patch into the corresponding memory location. Together, these two steps form a stable and interpretable update mechanism, ensuring that Gated DeltaNet both preserves useful long-range context and selectively incorporates new evidence.

These three equations can be equivalently compacted into a single matrix form:
\begin{equation}
S_t = S_{t-1}(\alpha_t (I - \beta_t k_t k_t^\top)) + \beta_t v_t k_t^\top,
\end{equation}
which simultaneously attenuates the old memory and incorporates the new information selectively and efficiently.

The chunkwise GDN update used in our framework is expressed as
\begin{equation}
H_{\text{global\_seq}} = \mathrm{GDN}(Q, K, V; \alpha, \beta),
\end{equation}
providing an efficient and expressive global representation for extremely long WSI patch sequences.

The local branch output sequence $Z_{\text{local\_seq}}$ is first passed through a linear projection to produce $H_{\text{local\_seq}}$. The global output $H_{\text{global\_seq}}$ and local branch output $H_{\text{local\_seq}}$  
are fused with a learnable gate:

\vspace{-10pt}\begin{align}
G &= \sigma(W_g H),\\
O &= G \odot H_{\text{global\_seq}} + (1 - G) \odot H_{\text{local\_seq}}.
\end{align}\vspace{-20pt}

\begin{table}[t]
\caption{Overview of datasets and corresponding tasks.}
\label{tab:tcga_survival}
\tablestyle{22pt}{1.0}
\centering
\footnotesize
\begin{tabular}{lcc}
\toprule
Dataset & Tasks & \#WSIs \\
\midrule
BLCA     & Survival & 437 \\
BRCA     & Survival & 1023 \\
COADREAD & Survival & 572 \\
KIRC     & Survival & 498 \\
KIRP     & Survival & 261 \\
LUAD     & Survival & 455 \\
STAD     & Survival & 363 \\
UCEC     & Survival & 539 \\
BRACS    & Classification & 545 \\
NSCLC    & Classification & 1053 \\
\bottomrule
\end{tabular}\vspace{-7pt}
\end{table}

\begin{table*}[t]
\tablestyle{6.5pt}{1.1}
\centering
\footnotesize  
\caption{Performance comparison across eight TCGA~\cite{tcga} survival prediction datasets using the ResNet-50~\cite{resnet} feature extractor.}
\label{tab:survival_resnet}
\begin{tabular}{l c c c c c c c c}
\toprule
\diagbox{Method}{Dataset} &
\multicolumn{1}{c}{BLCA} & \multicolumn{1}{c}{BRCA} & \multicolumn{1}{c}{COADREAD} & \multicolumn{1}{c}{KIRC} & \multicolumn{1}{c}{KIRP} & \multicolumn{1}{c}{LUAD} & \multicolumn{1}{c}{STAD} & \multicolumn{1}{c}{UCEC}\\
\midrule 
\textbf{Max-Pooling}  & {0.5555\tiny$\pm$0.0797} & {0.5462\tiny$\pm$0.0791} & {0.5638\tiny$\pm$0.0448} & {0.6235\tiny$\pm$0.0724} & {0.5826\tiny$\pm$0.1158} & {0.6180\tiny$\pm$0.0417} & {0.5389\tiny$\pm$0.0373} & {0.6397\tiny$\pm$0.1190} \\
\textbf{Mean-Pooling} & {0.5904\tiny$\pm$0.0494} & {0.5827\tiny$\pm$0.0651} & {0.5781\tiny$\pm$0.0457} & {0.6334\tiny$\pm$0.0514} & {0.6619\tiny$\pm$0.0297} & {0.6186\tiny$\pm$0.0545} & {0.5906\tiny$\pm$0.0470} & {0.6994\tiny$\pm$0.0889} \\
\textbf{ABMIL}~\cite{abmil}  & {0.5679\tiny$\pm$0.0342} & {0.5742\tiny$\pm$0.0924} & {0.5596\tiny$\pm$0.0614} & {0.6737\tiny$\pm$0.0536} & {0.7028\tiny$\pm$0.0689} & {0.6095\tiny$\pm$0.0621} & {0.5671\tiny$\pm$0.0528} & {0.6990\tiny$\pm$0.0990}  \\
\textbf{CLAM}~\cite{clam} & {0.5800\tiny$\pm$0.0132} & {0.6142\tiny$\pm$0.0182} & {0.6079\tiny$\pm$0.0199} & {0.5953\tiny$\pm$0.0112} & {0.7099\tiny$\pm$0.0614} & {0.5941\tiny$\pm$0.0333} & {0.5873\tiny$\pm$0.0260} & {0.6343\tiny$\pm$0.0459}\\
\textbf{DSMI}L~\cite{dsmil} & {0.5991\tiny$\pm$0.0203} & {0.6223\tiny$\pm$0.0450} & {0.6087\tiny$\pm$0.0214} & {0.5956\tiny$\pm$0.0225} & {0.7380\tiny$\pm$0.0504} & {0.6135\tiny$\pm$0.0303} & {0.5904\tiny$\pm$0.0283} & {0.6603\tiny$\pm$0.0356}\\
\textbf{TransMIL}~\cite{transmil} & {0.6475\tiny$\pm$0.0322} & {0.5813\tiny$\pm$0.0827} & {0.5964\tiny$\pm$0.0445} & {0.6339\tiny$\pm$0.0473} & {0.6815\tiny$\pm$0.1004} & {0.6214\tiny$\pm$0.0423} & {0.6047\tiny$\pm$0.0393} & {0.7155\tiny$\pm$0.0942}  \\
\textbf{S4MIL}~\cite{s4mil} & \underline{0.6501\tiny$\pm$0.0356} & {0.6248\tiny$\pm$0.0576} & {0.6545\tiny$\pm$0.0743} & {0.6696\tiny$\pm$0.0350} & {0.7257\tiny$\pm$0.0391} & {0.6413\tiny$\pm$0.0470} & {0.6081\tiny$\pm$0.0055} & {0.7452\tiny$\pm$0.1028} \\
\textbf{MambaMIL}~\cite{mambamil} & {0.6398\tiny$\pm$0.0376} & \underline{0.6364\tiny$\pm$0.0524} & {0.6462\tiny$\pm$0.0678} & {0.6958\tiny$\pm$0.0573} & {0.7070\tiny$\pm$0.0534} & {0.6396\tiny$\pm$0.0382} & \underline{0.6340\tiny$\pm$0.0282} & \underline{0.7526\tiny$\pm$0.0717}  \\
\textbf{Mamba2MIL}~\cite{mamba2mil} & {0.6317\tiny$\pm$0.0357} & {0.6254\tiny$\pm$0.0542} & \underline{0.6651\tiny$\pm$0.0242} & \underline{0.6976\tiny$\pm$0.0503} & \underline{0.7473\tiny$\pm$0.0719} & \underline{0.6521\tiny$\pm$0.0533} & {0.6207\tiny$\pm$0.0464} & {0.7524\tiny$\pm$0.0975} \\
\rowcolor{gray!20}
\textbf{\thename{}(ours)} & \textbf{0.6604\tiny$\pm$0.0418} & \textbf{0.6593\tiny$\pm$0.0428} & \textbf{0.6857\tiny$\pm$0.0634} & \textbf{0.7166\tiny$\pm$0.0385} & \textbf{0.8016\tiny$\pm$0.0600} & \textbf{0.6561\tiny$\pm$0.0695} & \textbf{0.6403\tiny$\pm$0.0590} & \textbf{0.7712\tiny$\pm$0.0932}\\
\bottomrule
\end{tabular}
\end{table*}

\begin{table*}[t]
\tablestyle{6.5pt}{1.1}
\centering
\footnotesize  
\caption{Performance comparison across eight TCGA~\cite{tcga} survival prediction datasets using the UNI~\cite{uni} feature extractor.}
\label{tab:survival_uni}
\begin{tabular}{l c c c c c c c c}
\toprule
\diagbox{Method}{Dataset} &
\multicolumn{1}{c}{BLCA} & \multicolumn{1}{c}{BRCA} & \multicolumn{1}{c}{COADREAD} & \multicolumn{1}{c}{KIRC} & \multicolumn{1}{c}{KIRP} & \multicolumn{1}{c}{LUAD} & \multicolumn{1}{c}{STAD} & \multicolumn{1}{c}{UCEC}\\
\midrule 
\textbf{Max-Pooling}  & {0.6301\tiny$\pm$0.0694} & {0.5608\tiny$\pm$0.0821} & {0.6153\tiny$\pm$0.0505} & {0.6751\tiny$\pm$0.0433} & {0.5662\tiny$\pm$0.1621} & {0.6007\tiny$\pm$0.0407} & {0.5369\tiny$\pm$0.0523} & {0.6541\tiny$\pm$0.0887} \\
\textbf{Mean-Pooling} & {0.6524\tiny$\pm$0.0273} & \underline{0.6776\tiny$\pm$0.0453} & {0.6650\tiny$\pm$0.0603} & {0.7187\tiny$\pm$0.0443} & {0.7913\tiny$\pm$0.0252} & {0.6345\tiny$\pm$0.0625} & {0.6120\tiny$\pm$0.0610} & {0.7473\tiny$\pm$0.0929} \\
\textbf{ABMIL}~\cite{abmil} & \underline{0.6835\tiny$\pm$0.0426} & {0.6490\tiny$\pm$0.0499} & \textbf{0.6875\tiny$\pm$0.0520} & \underline{0.7233\tiny$\pm$0.0385} & {0.7773\tiny$\pm$0.0578} & \underline{0.6540\tiny$\pm$0.0604} & {0.6142\tiny$\pm$0.0647} & \underline{0.7702\tiny$\pm$0.0866} \\
\textbf{CLAM}~\cite{clam} & {0.6039\tiny$\pm$0.0191} & {0.6401\tiny$\pm$0.0337} & {0.6276\tiny$\pm$0.0234} & {0.5937\tiny$\pm$0.0126} & {0.7518\tiny$\pm$0.0463} & {0.5933\tiny$\pm$0.0260} & {0.5963\tiny$\pm$0.0298} & {0.6233\tiny$\pm$0.0226} \\
\textbf{DSMIL}~\cite{dsmil} & {0.5875\tiny$\pm$0.0182} & {0.6155\tiny$\pm$0.0307} & {0.6518\tiny$\pm$0.0477} & {0.5994\tiny$\pm$0.0253} & {0.6990\tiny$\pm$0.0441} & {0.5992\tiny$\pm$0.0204} & {0.6120\tiny$\pm$0.0228} & {0.6412\tiny$\pm$0.0334} \\
\textbf{TransMIL}~\cite{transmil} & {0.6691\tiny$\pm$0.0362} & {0.6345\tiny$\pm$0.0511} & {0.6011\tiny$\pm$0.0457} & {0.6945\tiny$\pm$0.0464} & {0.6727\tiny$\pm$0.0648} & {0.6095\tiny$\pm$0.0542} & {0.5738\tiny$\pm$0.0783} & {0.7309\tiny$\pm$0.1029} \\
\textbf{S4MIL}~\cite{s4mil} & {0.6572\tiny$\pm$0.0355} & {0.6709\tiny$\pm$0.0551} & {0.6699\tiny$\pm$0.0637} & {0.7171\tiny$\pm$0.0421} & {0.7854\tiny$\pm$0.0519} & {0.6293\tiny$\pm$0.0584} & {0.6136\tiny$\pm$0.0572} & {0.7268\tiny$\pm$0.1212}  \\
\textbf{MambaMIL}~\cite{mambamil} & {0.6444\tiny$\pm$0.0546} & {0.6387\tiny$\pm$0.0583} & {0.6174\tiny$\pm$0.0625} & {0.7143\tiny$\pm$0.0480} & \underline{0.7978\tiny$\pm$0.0599} & {0.6086\tiny$\pm$0.0586} & {0.5914\tiny$\pm$0.0568} & {0.7296\tiny$\pm$0.0892} \\
\textbf{Mamba2MIL}~\cite{mamba2mil} & {0.6546\tiny$\pm$0.0464} & {0.6114\tiny$\pm$0.0592} & {0.6550\tiny$\pm$0.0585} & {0.7119\tiny$\pm$0.0498} & {0.7153\tiny$\pm$0.1160} & {0.6147\tiny$\pm$0.0539} & \underline{0.6237\tiny$\pm$0.0777} & {0.7325\tiny$\pm$0.0911} \\
\rowcolor{gray!20}
\textbf{\thename{}(ours)} & \textbf{0.6885\tiny$\pm$0.0303} & \textbf{0.6898\tiny$\pm$0.0589} & \underline{0.6825\tiny$\pm$0.0683} & \textbf{0.7273\tiny$\pm$0.0607} & \textbf{0.8312\tiny$\pm$0.0337} & \textbf{0.6560\tiny$\pm$0.0461} & \textbf{0.6441\tiny$\pm$0.0353} & \textbf{0.7713\tiny$\pm$0.0980} \\
\bottomrule
\end{tabular}
\end{table*}

\subsection{Architecture Details}
The \thename{} model consists of a single Gated DeltaNet layer designed to capture both local and global dependencies within the patch sequence. This layer employs 6 attention heads to enable multi-head interactions among patch tokens, with each token represented as a 128-dimensional embedding.

Across our experiments on most datasets, slide-level feature normalization is generally not required.
However, for the BLCA dataset, we apply z-score normalization to each patch feature before feeding them into the model. This normalization stabilizes feature statistics and can improve convergence and performance for this dataset, while for other datasets, the original features are used directly.

\section{EXPERIMENTS}
\subsection{Datasets}
To comprehensively evaluate the proposed \thename{} framework, we conduct experiments on two categories of benchmarks: (1) survival analysis datasets and (2) slide-level classification datasets. The specific number of samples in each dataset is summarized in Table~\ref{tab:tcga_survival}.

{\bf Survival datasets.} For survival prediction experiments, we utilize several TCGA~\cite{tcga} cohorts, including BLCA, BRCA, COADREAD, KIRC, KIRP, LUAD, STAD, and UCEC. All datasets consist of whole-slide images annotated with patient survival information. To ensure robust evaluation, we adopt a 5-fold cross-validation strategy for all cohorts.

{\bf Classification datasets.} We conduct experiments on two widely used computational pathology classification datasets: BRACS~\cite{bracs} and NSCLC.

The BRACS~\cite{bracs} dataset contains breast tissue WSIs spanning seven categories, including six lesion subtypes, \textit{i.e.}, Pathological Benign (PB), Invasive Carcinoma (IC), Ductal Carcinoma in Situ (DCIS), Atypical Ductal Hyperplasia (ADH), Flat Epithelial Atypia (FEA), Usual Ductal Hyperplasia (UDH), and normal tissue samples. We adopt 10-fold cross-validation for evaluation.

The NSCLC dataset~\cite{tcga} comprises 541 lung adenocarcinoma (LUAD) cases and 512 lung squamous cell carcinoma (LUSC) cases, representing the two major histological subtypes of non-small cell lung cancer.
For each dataset, we adopt 10-fold cross-validation for evaluation.

\begin{table*}[t]
\centering
\footnotesize  
\tablestyle{16pt}{1.1}
\caption{Performance comparison on BRACS~\cite{bracs} and NSCLC~\cite{tcga} classification using the ResNet-50~\cite{resnet} feature extractor.}
\label{tab:classification_renset}
\begin{tabular}{l c c c c c c}
\toprule
\diagbox{Method}{Dataset} &
\multicolumn{2}{c}{BRACS} & 
\multicolumn{2}{c}{NSCLC} &
\multicolumn{2}{c}{MEAN} \\
\cmidrule(lr){2-3} \cmidrule(lr){4-5} \cmidrule(lr){6-7}  
 & AUC & ACC  & AUC & ACC & AUC & ACC \\
\midrule 
Max-Pooling  & {0.7428\tiny$\pm$0.0514} & {0.4368\tiny$\pm$0.0581}  & {0.9492\tiny$\pm$0.0156} & {0.8780\tiny$\pm$0.0375} & 0.8460 & 0.6574 \\
Mean-Pooling & {0.7426\tiny$\pm$0.0338} & {0.4293\tiny$\pm$0.0545}  & {0.9386\tiny$\pm$0.0236} & {0.8619\tiny$\pm$0.0324} & 0.8406 & 0.6456 \\
ABMIL~\cite{abmil} & {0.7498\tiny$\pm$0.0430} & {0.4106\tiny$\pm$0.0572}  & {0.9346\tiny$\pm$0.0245} & {0.8685\tiny$\pm$0.0290} & 0.8422 & 0.6396 \\
CLAM~\cite{clam} & {0.7449\tiny$\pm$0.0437} & {0.4162\tiny$\pm$0.0584}  & {0.9426\tiny$\pm$0.0246} & {0.8561\tiny$\pm$0.0328} & 0.8438 & 0.6362 \\
DSMIL~\cite{dsmil} & {0.7805\tiny$\pm$0.0460} & {0.4572\tiny$\pm$0.0601}  & \underline{0.9541\tiny$\pm$0.0144} & {0.8780\tiny$\pm$0.0257} & 0.8673 & 0.6676 \\
TransMIL~\cite{transmil} & {0.7099\tiny$\pm$0.0699} & {0.3531\tiny$\pm$0.0751}  & {0.9221\tiny$\pm$0.0160} & {0.8342\tiny$\pm$0.0342} & 0.8160 & 0.5937 \\
S4MIL~\cite{s4mil} & {0.7763\tiny$\pm$0.0508} & {0.4517\tiny$\pm$0.0706}  & {0.9232\tiny$\pm$0.0291} & {0.8333\tiny$\pm$0.0481} & 0.8498 & 0.6425 \\
MambaMIL~\cite{mambamil} & {0.7808\tiny$\pm$0.0453} & {0.4161\tiny$\pm$0.0760}  & {0.9430\tiny$\pm$0.0156} & {0.8733\tiny$\pm$0.0220} & 0.8619 & 0.6447 \\
Mamba2MIL~\cite{mamba2mil} & \underline{0.7924\tiny$\pm$0.0468} & \underline{0.4944\tiny$\pm$0.0390}  & {0.9524\tiny$\pm$0.0182} & \underline{0.8847\tiny$\pm$0.0267} & \underline{0.8724} & \underline{0.6896}\\
\rowcolor{gray!20}
\thename{} (ours) & \textbf{0.8208\tiny$\pm$0.0181} & \textbf{0.5313\tiny$\pm$0.0528}  & \textbf{0.9551\tiny$\pm$0.0170} & \textbf{0.8904\tiny$\pm$0.0246} & \textbf{0.8880} & \textbf{0.7109}  \\
\bottomrule
\end{tabular}\vspace{-5pt}
\end{table*}

\begin{table*}[t]
\centering
\footnotesize  
\tablestyle{16pt}{1.1}
\caption{Performance comparison on BRACS~\cite{bracs} and NSCLC~\cite{tcga} classification using the UNI~\cite{uni} feature extractor.}
\label{tab:classification_uni}
\begin{tabular}{l c c c c c c}
\toprule
\diagbox{Method}{Dataset} &
\multicolumn{2}{c}{BRACS} & 
\multicolumn{2}{c}{NSCLC} &
\multicolumn{2}{c}{MEAN} \\
\cmidrule(lr){2-3} \cmidrule(lr){4-5} \cmidrule(lr){6-7}  
 & AUC & ACC  & AUC & ACC & AUC & ACC \\
\midrule 
Max-Pooling  & {0.8067\tiny$\pm$0.0297} & {0.5371\tiny$\pm$0.0464} & {0.9443\tiny$\pm$0.0208} & {0.8771\tiny$\pm$0.0292} & 0.8755 & 0.7071 \\
Mean-Pooling & {0.8117\tiny$\pm$0.0292} & {0.5130\tiny$\pm$0.0368} & {0.9279\tiny$\pm$0.0212} & {0.8504\tiny$\pm$0.0229} & 0.8698 & 0.6817 \\
ABMIL~\cite{abmil} & {0.8400\tiny$\pm$0.0287} & {0.5558\tiny$\pm$0.0663} & {0.9344\tiny$\pm$0.0277} & {0.8704\tiny$\pm$0.0399} & 0.8872 & 0.7131 \\
CLAM~\cite{clam} & {0.8410\tiny$\pm$0.0233} & {0.5502\tiny$\pm$0.0418} & {0.9360\tiny$\pm$0.0315} & {0.8619\tiny$\pm$0.0351} & 0.8885 & 0.7061 \\
DSMIL~\cite{dsmil} & {0.8394\tiny$\pm$0.0263} & {0.5407\tiny$\pm$0.0537} & {0.9488\tiny$\pm$0.0150} & {0.8790\tiny$\pm$0.0250} & 0.8941 & 0.7099 \\
TransMIL~\cite{transmil} & {0.7566\tiny$\pm$0.0254} & {0.4683\tiny$\pm$0.0530} & {0.9233\tiny$\pm$0.0427} & {0.8400\tiny$\pm$0.0446} & 0.8400 & 0.6542 \\
S4MIL~\cite{s4mil} & \underline{0.8496\tiny$\pm$0.0328} & \underline{0.5707\tiny$\pm$0.0366} & {0.9165\tiny$\pm$0.0236} & {0.8361\tiny$\pm$0.0335} & 0.8831 & 0.7034 \\
MambaMIL~\cite{mambamil} & {0.8210\tiny$\pm$0.0436} & {0.5464\tiny$\pm$0.0679} & {0.9486\tiny$\pm$0.0188} & {0.8695\tiny$\pm$0.0301} & 0.8848 & 0.7080 \\
Mamba2MIL~\cite{mamba2mil} & {0.8360\tiny$\pm$0.0207} & {0.5595\tiny$\pm$0.0505} & \underline{0.9553\tiny$\pm$0.0189} & \underline{0.8847\tiny$\pm$0.0182} & \underline{0.8957} & \underline{0.7221} \\
\rowcolor{gray!20}
\thename{} (ours) & \textbf{0.8701\tiny$\pm$0.0349} & \textbf{0.6060\tiny$\pm$0.0516}  & \textbf{0.9580\tiny$\pm$0.0163} & \textbf{0.8923\tiny$\pm$0.0284} & \textbf{0.9141} & \textbf{0.7492}   \\
\bottomrule
\end{tabular}
\end{table*}

\subsection{Implementation Details}
For all datasets, we follow MambaMIL~\cite{mambamil} and extract 512×512 patches at ×20 magnification.
We conduct experiments using both ResNet-50~\cite{resnet} and UNI~\cite{uni} backbones with the feature dimension set to 1024.
The ResNet-50~\cite{resnet} model is pre-trained on ImageNet~\cite{imagenet}, whereas the UNI~\cite{uni} model is pre-trained on more than 100,000 diagnostically annotated H\&E-stained WSIs.
The full training configurations are listed in Appendix~\ref{sec:train_cfg}.
For classification tasks, we report accuracy (ACC) and area under the ROC curve (AUC), while for survival prediction tasks, the concordance index (C-index) is used.
For visualization, we use a 2-layer configuration. Although we observed that the 1-layer variant achieves slightly better performance, we keep the 2-layer setup for consistency and interpretability in the visualization process.
The details on the compared methods are listed in appendix~\ref{sec:comp_methods}.

\subsection{Results}
In this section, we present the main results of our method in comparison with existing methods.
Fig.~\ref{fig:radar} presents a brief comparison of various MIL methods across six key evaluation metrics, including survival prediction C-index, AUC, and ACC on both ResNet-50~\cite{resnet} and UNI~\cite{uni} feature extractors.
Our method provides the most extensive and balanced coverage across all six metrics and shows consistent improvements across various tasks.
The gains are particularly pronounced on the UNI-based~\cite{uni} metrics, which reflect the robustness and strong generalization ability of our framework.
The detailed results are analyzed below.

\subsubsection{Survival Prediction}
Table~\ref{tab:survival_resnet} and Table~\ref{tab:survival_uni} report the performance of multiple MIL approaches across eight TCGA~\cite{tcga} survival datasets using ResNet-50~\cite{resnet} and UNI~\cite{uni} feature extractors. Overall, our method \thename{} consistently achieves the leading results under both settings.
This result demonstrates strong generalization ability and robustness of \thename{} across feature extractor types.

As illustrated in Table~\ref{tab:survival_resnet}, with ResNet-50~\cite{resnet} features, our method \thename{} achieves the highest average C-index of 0.6989, and outperforms the strongest baseline Mamba2MIL~\cite{mamba2mil} by 3.69\%. 
This demonstrates that \thename{} remains highly effective even when paired with a relatively simple feature extractor.

Furthermore, as presented in Table~\ref{tab:survival_uni}, using UNI~\cite{uni} features leads to an overall performance increase for all methods, which confirms the benefit of pathology-specific pretraining for survival prediction.
Under this stronger feature representation, \thename{} continues to outperform all baselines and achieves a mean C-index of 0.7113, which represents a 2.36\% improvement over the strongest competitor and ranks first on seven of the eight datasets, demonstrating robust generalization across diverse data types.

\subsubsection{Classification}
As shown in Table~\ref{tab:classification_renset} and Table~\ref{tab:classification_uni}, \thename{} delivers the most stable and superior performance across datasets and backbones.

Using ResNet-50~\cite{resnet} features (Table~\ref{tab:classification_renset}), our method achieves the strongest performance among all compared MIL approaches on both BRACS and NSCLC. 
When averaged across the two datasets, our method obtains an AUC of 0.8880 and an ACC of 0.7109, with average improvements of 1.79\% in AUC and 3.09\% in ACC, representing the best overall performance under the ResNet-50~\cite{resnet} setting.

As illustrated in Table~\ref{tab:classification_uni}, with UNI~\cite{uni} features, all methods improve due to stronger pretraining, yet our method maintains a consistent and clear leading advantage. 
Our approach reaches an average AUC of 0.9141 and an average ACC of 0.7492, with improvements of 2.05\% in AUC and 3.75\% in ACC across both datasets.

These results collectively indicate that our model benefits more effectively from both backbone types and consistently generalizes better than existing MIL architectures.

\subsection{Ablation Study}

To validate the effectiveness of each module, we perform ablation studies on the BRACS dataset~\cite{bracs}, using a ResNet-50~\cite{resnet} backbone as the feature extractor.

\subsubsection{Model Design}
Table~\ref{tab:ablation_model} presents the ablation study of our model design.
The results highlight the contributions of each component in our model.
When the local module is removed, the model’s performance decreases to an AUC of 0.8013 and an ACC of 0.5073 compared to the full model.
This suggests that incorporating local information enables the model to capture fine-grained pathological features better,
thereby improving both classification accuracy and discriminative ability.
Removing the gated module also reduces performance, with an AUC of 0.7921 and ACC of 0.4924. This suggests that the gating mechanism is important for allowing the model to focus on key instances while suppressing irrelevant or noisy regions.
Excluding the delta module results in a performance drop, with an AUC of 0.8172 and an ACC of 0.5110, demonstrating that the delta component contributes to capturing discriminative features across instances.
The full model, incorporating all three components, achieves the best performance, confirming that the local, gated, and delta modules are complementary and lead to optimal results.

\begin{table}[t]
\centering
\tablestyle{15pt}{1.3}
\caption{Ablation study on the model design.}
\label{tab:ablation_model}
\begin{tabular}{ccccc}
\toprule
 Local & Gated & Delta & AUC & ACC \\
\midrule
 \xmark & \cmark & \cmark & 0.8013 & 0.5073  \\
 \cmark & \xmark & \cmark & 0.7921 & 0.4924  \\
 \cmark & \cmark & \xmark & 0.8172 & 0.5110  \\
\rowcolor{gray!20}
 \cmark & \cmark & \cmark & \textbf{0.8208} & \textbf{0.5313}  \\
\bottomrule
\end{tabular}
\end{table}

\subsubsection{Layer Number}
In Table~\ref{tab:ablation_layer}, we conduct an ablation study on the number of layers to evaluate how model depth influences the performance.
The results show that the 1-layer architecture achieves the highest AUC and ACC, while increasing the number of layers to 2 or 3 leads to a consistent performance drop.
This indicates that a shallow structure is sufficient to capture the key bag-level dependencies, as the weak supervision and heterogeneity make deeper recurrent stacking more sensitive to noise and over-smoothing of discriminative patches.
Moreover, deeper configurations are harder to optimize under limited supervision and lead to diminished generalization.

\begin{table}[t]
\centering
\footnotesize
\tablestyle{30pt}{1.3}
\caption{Ablation study on the number of layers.}
\label{tab:ablation_layer}
\begin{tabular}{ccc}
\toprule
{Layers} & AUC & ACC \\
\midrule
\rowcolor{gray!20}
1   & \textbf{0.8208} & \textbf{0.5313} \\
2   & 0.7849 & 0.5201 \\
3   & 0.7896 & 0.4963 \\
\bottomrule
\end{tabular}
\end{table}

\section{Discussion}
\subsection{Principal Findings and Mechanism Efficacy}

\begin{figure}[t]
\centering
\includegraphics[width=\columnwidth]{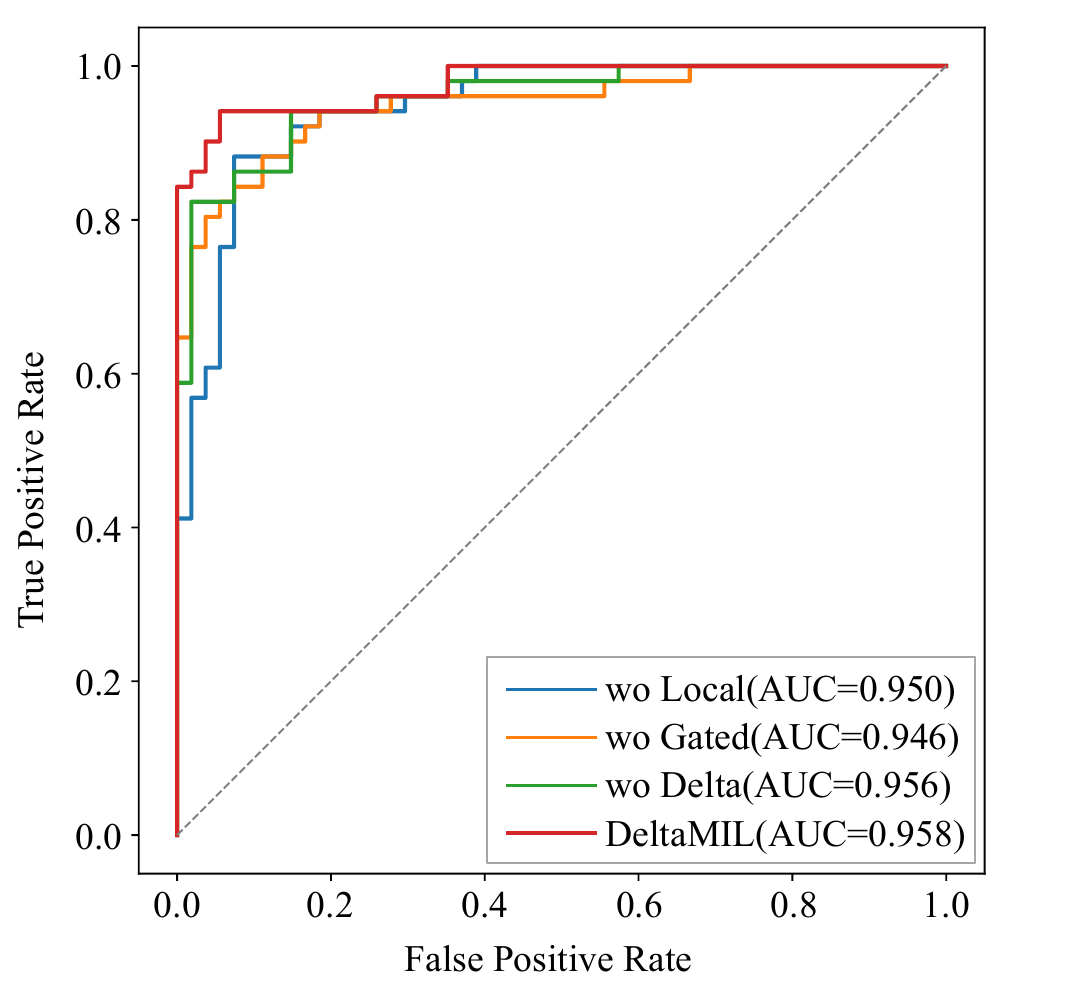}
\caption{Comparison of ROC curves for different model variants in the ablation study on the first fold of the NSCLC~\cite{tcga} test set.}
\label{fig:roc}
\end{figure}

The primary challenge in WSI analysis lies in the massive scale of the images and the large proportion of background or uninformative regions, which can make it difficult for models to identify and leverage the small fraction of diagnostically relevant patches.
Our proposed \thename{} framework addresses these issues by dynamically integrating and filtering the most relevant features.

As shown in Figure~\ref{fig:roc}, we present the ROC curves for different model variants in the ablation study on the NSCLC dataset using UNI feature~\cite{uni}.
The delta mechanism effectively eliminates non-discriminative memory content while selectively incorporating highly relevant information that is strongly correlated with the current patch.
The gating module rapidly removes irrelevant signals such as background tissue or artifacts.
The local pattern mixing preserves fine-grained pathological structures.
Together, these components enable \thename{} to achieve robust ROC performance and high AUC scores, which demonstrate that each mechanism contributes meaningfully to discriminative feature extraction and overall classification accuracy in WSI analysis.
This combination ensures critical global context and fine-grained local details are preserved for reliable predictions.

\begin{figure*}[!t]
\centering
\includegraphics[width=\textwidth]{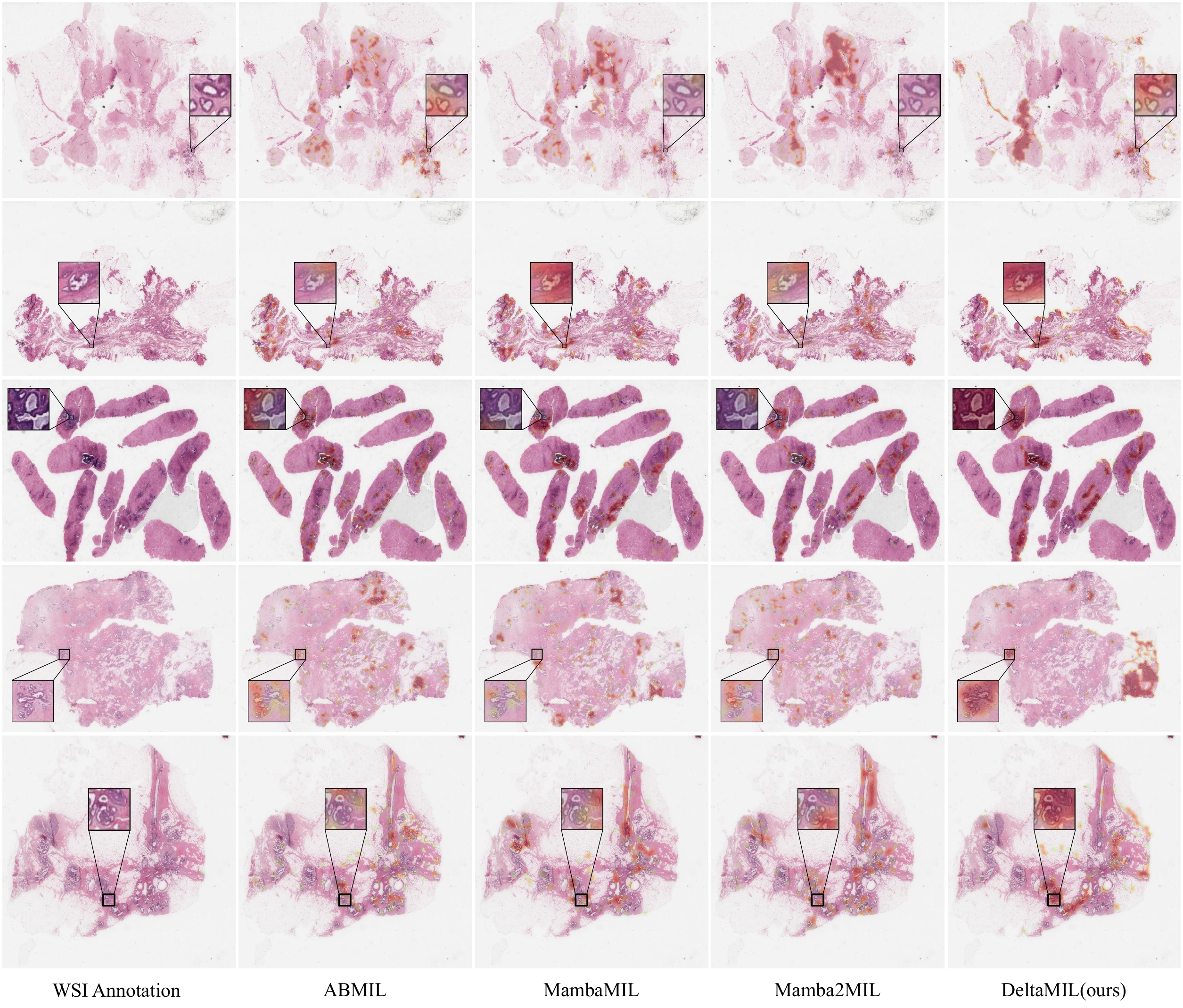}
\caption{Visualization of attention regions generated by different methods for a WSI from the BRACS dataset~\cite{bracs}. The first column shows the expert annotation of the lesson regions, while the remaining columns present the attention visualizations produced by each method. For consistent comparison, each method displays the same number of high-attention regions. Darker colors indicate higher attention scores.}
\label{fig:visualization}
\end{figure*}

\subsection{Interpretability and Potential Clinical Utility}
Beyond quantitative metrics, the interpretability of predictions in computational pathology is essential for supporting informed clinical decision-making.
To more intuitively demonstrate the attention distribution of \thename{} and its advantages in identifying key regions, we visualize the activation maps produced by several representative methods on the BRACS dataset~\cite{bracs}.
As illustrated in Fig.~\ref{fig:visualization}, we visualize the activation maps of an attention-based method (ABMIL~\cite{abmil}) and two SSM-based methods (MambaMIL~\cite{mambamil} and Mamba2MIL~\cite{mamba2mil}).
For a fair comparison, we visualize the same number of top-attended regions for each method. 
The visualizations show that \thename{} strongly focuses on diagnostically significant pathological regions while exhibiting minimal responses in non-critical areas. These results provide compelling evidence that the model accurately and consistently captures discriminative pathological features and offers improved interpretability.

The proposed \thename{} demonstrates strong potential for clinical application in computational pathology. By accurately identifying diagnostically relevant regions and providing interpretable attention visualizations, the model can assist pathologists in focusing on critical pathological structures during slide review. Such region-level interpretability may help reduce diagnostic uncertainty, improve consistency across observers, and support a more efficient workflow by prioritizing high-risk regions for further examination.

\subsection{Limitations and Future Work}
\subsubsection{Limitations}
Our method has been evaluated primarily on WSI classification and survival prediction tasks, and its effectiveness on a broader range of medical applications remains to be validated.
Moreover, the reliance on pre-extracted patch features implies that the upstream feature extractor acts as a performance bottleneck, directly determining the fidelity of the downstream representations.

\subsubsection{Future work}
To address these limitations, future work will focus on developing an end-to-end framework that jointly optimizes feature extraction and aggregation to improve the discriminative quality of learned representations.
Furthermore, the broader applicability will be investigated across diverse clinical scenarios and multi-granularity pathological tasks, \emph{e.g.}, segmentation, to extend \thename{} beyond classification and survival prediction to more complex downstream analyses.

\section{Conclusion}
In this work, we proposed \thename{}, a new MIL framework designed to more effectively identify and integrate discriminative information from large and heterogeneous WSIs. By combining gated filtering, delta-based selective updating, and adaptive local pattern integration, \thename{} strengthens the model’s ability to suppress redundant signals while preserving and accumulating meaningful pathological cues. This design enables the model to maintain robust focus on semantically relevant regions and ensures that fine-grained local details contribute to the global slide representation. Extensive experiments on both survival prediction and slide-level classification demonstrate that \thename{} consistently outperforms existing MIL approaches under different feature extractors, highlighting its strong generalizability and reliability across diverse WSI analysis tasks.

\appendices

\section{Training configurations}
\label{sec:train_cfg}
In all experiments, we follow the general training protocol of MambaMIL~\cite{mambamil}.
Specifically, we use the Adam optimizer~\cite{adam}.
The Adam parameters are set as $\beta_1=0.9,\beta_2=0.999,\epsilon=1\times 10^{-8}$.
The weight decay is set to $1\times10^{-5}$.
We set the batch size to 1.
For survival prediction tasks, we leverage dropout with a dropout rate of 0.25.
We perform gradient accumulation over 32 steps.
For classification tasks, dropout and gradient accumulation are not used.
The learning rates are independently tuned for each dataset and feature extractor, as shown in Table~\ref{tab:learning_rate}.
Early stopping is applied based on the validation metric to prevent overfitting.
All experiments are performed using an NVIDIA RTX 3090 GPU.

\begin{table}[ht]
\centering
\footnotesize
\tablestyle{20pt}{1.0}
\caption{Learning rate settings for each dataset.}
\label{tab:learning_rate}
\begin{tabular}{ccc}
\toprule
Dataset & ResNet-50~\cite{resnet} & UNI~\cite{uni} \\
\midrule
BLCA                & 1e-4 & 2e-4 \\
BRCA                & 5e-5 & 5e-5 \\
COADREAD            & 1e-4 & 2e-4 \\
KIRC                & 1e-4 & 3e-4 \\
KIRP                & 2e-4 & 3e-4 \\
LUAD                & 6e-5 & 2e-4 \\
STAD                & 1e-4 & 3e-4 \\
UCEC                & 2e-4 & 4e-4 \\
BRACS~\cite{bracs}  & 1e-5& 5e-5 \\
NSCLC               & 5e-5 & 5e-5 \\
\bottomrule
\end{tabular}
\end{table}

\section{Details on the compared Methods}
\label{sec:comp_methods}
We benchmark our model against representative MIL paradigms spanning from classical to modern designs. Simple aggregation baselines such as Max-Pooling and Mean-Pooling provide a reference for non-parametric instance fusion. To capture instance importance, we include attention-based models such as ABMIL~\cite{abmil}, CLAM~\cite{clam}, DSMIL~\cite{dsmil}, and TransMIL~\cite{transmil}, as well as state-space–driven architectures like S4MIL~\cite{s4mil}, MambaMIL~\cite{mambamil}, and Mamba2MIL~\cite{mamba2mil}.
Together, these methods form a comprehensive benchmark spanning the major families of MIL designs.

Max-Pooling and Mean-Pooling serve as simple aggregation baselines, either selecting the most activated patches or treating all patches equally. ABMIL~\cite{abmil} introduces permutation-invariant attention to weigh patches according to their contribution, providing both interpretability and improved predictions. CLAM~\cite{clam} refines this approach by applying clustering constraints on representative patches, enabling the identification of diagnostically relevant regions without spatial annotations. DSMIL~\cite{dsmil} employs a dual-stream aggregator with self-supervised contrastive learning and multiscale feature fusion to enhance patch representations and slide-level predictions. TransMIL~\cite{transmil} leverages a Transformer-based MIL architecture to capture correlations among patches while integrating morphological and spatial information. S4MIL~\cite{s4mil} utilizes structured state space models to compress long patch sequences into memory units while preserving long-range dependencies. MambaMIL~\cite{mambamil} adopts the Mamba sequential model with sequence reordering to capture comprehensive instance relationships efficiently. Mamba2MIL~\cite{mamba2mil} extends this approach by incorporating weighted feature selection and sequence transformation, enabling more effective feature fusion and local sequence preservation.

\section*{References}
\bibliographystyle{ieeetr}
\bibliography{main}

\end{document}